\documentclass[letterpaper, 10 pt, conference]{ieeeconf}  

\IEEEoverridecommandlockouts                              

\title{\LARGE \bf
OmniSLAM: Omnidirectional Localization and Dense Mapping \\for Wide-baseline Multi-camera Systems
}

\author{Changhee Won$^{1*}$, Hochang Seok$^{1*}$, Zhaopeng Cui$^2$, Marc Pollefeys$^{2}$, and Jongwoo Lim$^{1\dagger}$
\thanks{$^*$Authors contributed equally.}
\thanks{$^\dagger$Corresponding author.}
\thanks{$^{1}$Department of Computer Sicence, Hanyang University, Korea.
        {\tt\small \{chwon, hochangseok, jlim\}@hanyang.ac.kr}}%
\thanks{$^{2}$Department of Computer Science, ETH Zurich,  Switzerland.
        {\tt\small \{zhaopeng.cui, marc.pollefeys\}@inf.ethz.ch}}%
}

\usepackage{gensymb}
\usepackage{graphicx}
\usepackage[font=small,labelfont=bf,singlelinecheck=false]{caption}
\usepackage[subrefformat=parens]{subcaption}
\usepackage{kotex}
\usepackage{bm}
\usepackage{amsmath}
\usepackage{amssymb}
\let\labelindent\relax
\usepackage{enumitem}
\usepackage{subcaption}
\usepackage{color}
\usepackage{booktabs}
\usepackage{multirow}
\usepackage{siunitx}
\usepackage{threeparttable}

\makeatletter
\let\NAT@parse\undefined
\makeatother

\usepackage{cite}
\usepackage{hyperref}
\hypersetup{colorlinks,linktocpage=true}

\newcommand{\tb}[1]{\textbf{#1}}
\newcommand{\mb}[1]{\mathbf{#1}}
\newcommand{\etal}[0]{\textit{et al.}}

\newcommand{\subsec}[1]{\vspace{4pt}{\setlength{\parindent}{0pt}\textbf{#1}~}}
\graphicspath{{./img/}}

\begin{document}

\maketitle
\thispagestyle{empty}
\pagestyle{empty}

\begin{abstract}

In this paper, we present an omnidirectional localization and dense mapping system for a wide-baseline multi-view stereo setup with ultra-wide field-of-view (FOV) fisheye cameras, which has a 360$\degree$ coverage of stereo observations of the environment.
For more practical and accurate reconstruction, we first introduce improved and light-weighted deep neural networks for the omnidirectional depth estimation, which are faster and more accurate than the existing networks.
Second, we integrate our omnidirectional depth estimates into the visual odometry (VO) and add a loop closing module for global consistency.
Using the estimated depth map, we reproject keypoints onto each other view, which leads to a better and more efficient feature matching process.
Finally, we fuse the omnidirectional depth maps and the estimated rig poses into the truncated signed distance function (TSDF) volume to acquire a 3D map.
We evaluate our method on synthetic datasets with ground-truth and real-world sequences of challenging environments, and the extensive experiments show that the proposed system generates excellent reconstruction results in both synthetic and real-world environments.

\end{abstract}

\section{INTRODUCTION}
3D geometric mapping of an environment is an essential part of autonomous navigation for cars or robots.
To this end, many range sensors, for example, laser-based LiDAR~\cite{ye2019tightly} and structured light 3D scanners~\cite{newcombe2011kinectfusion} can be used as their depth sensing is accurate, which is critical to the mapping results.
However, they often suffer from low vertical resolution, inter-sensor interference due to emitting light operation, or practical issues such as size, cost, and power consumption.
Meanwhile, the camera-based systems~\cite{geiger2010efficient,geiger2011stereoscan} are also used for the 3D dense mapping since they have a sufficient resolution with lower cost and smaller size, and the sensors operate passively.
Although cameras do not directly sense the distances from surfaces, 3D depth can be estimated by using a multi-view stereo.
Moreover, due to the drastic performance improvement recent deep learning-based algorithms for the stereo depth estimates have shown~\cite{zhang2019ga, chang2018pyramid, ilg2018occlusions}, the camera-based systems have become more favorable.

\begin{figure}[hbt!]
    \centering
    \includegraphics[width=\linewidth]{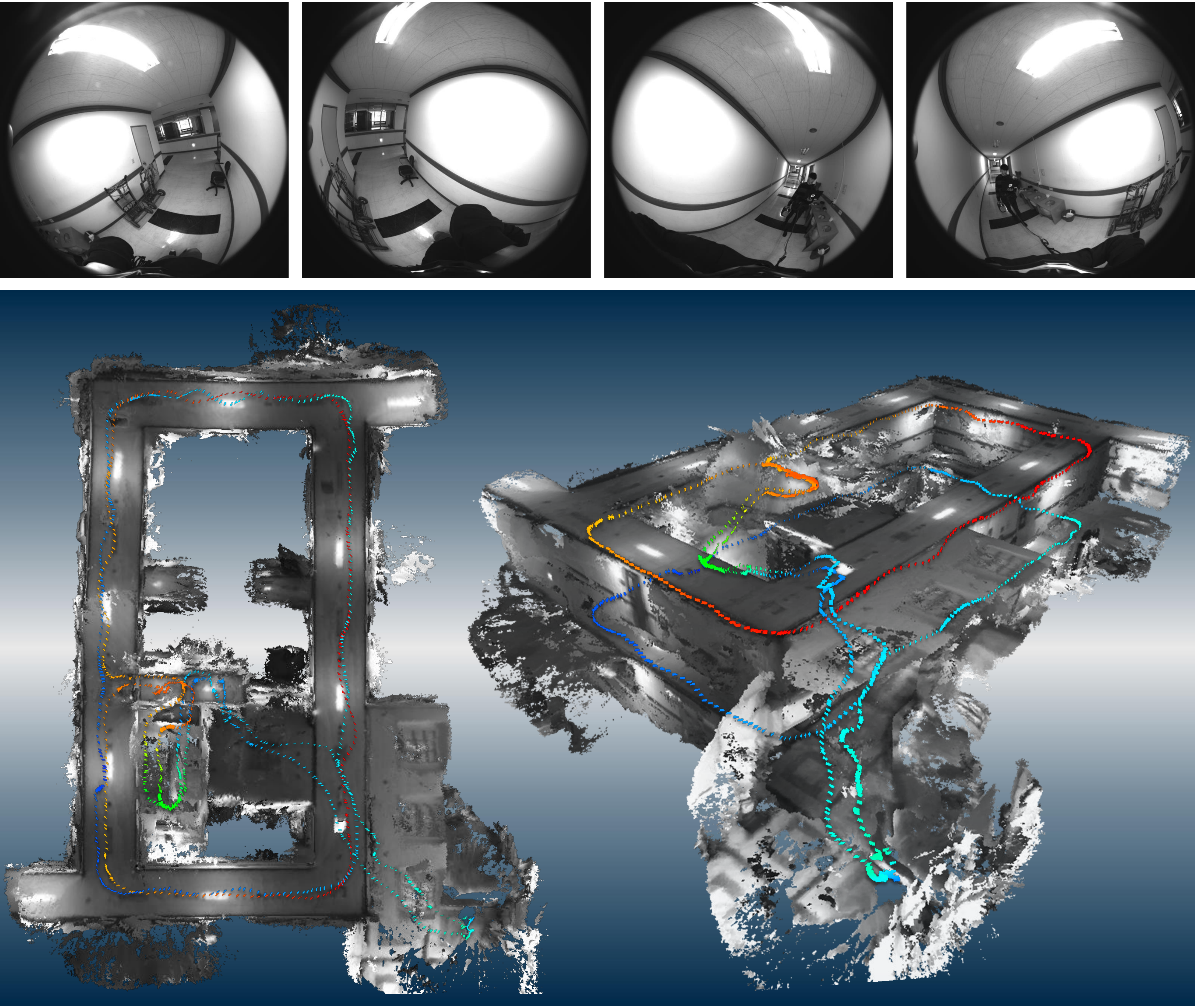}
    \caption{Top: example input images of a challenging indoor environment. Bottom: Densely reconstructed map of a duplex building with estimated trajectory. The colors of the trajectory encode heights.}
    \label{fig:intro}
\vspace{-15pt}
\end{figure}

Traditional stereo camera systems consist of two or more cameras looking in the same direction to find stereo correspondences from a set of rectified images.
After computing stereo disparities, corresponding output depth maps can be fused temporarily into a global map using camera poses which can be estimated using image features, for example, ORB feature descriptor~\cite{rublee2011orb,mur2015orb}.
For the dense mapping, both the depths and the camera poses need to be accurately estimated and are critical to the mapping results.

Meanwhile, there have also been strong needs on omnidirectional sensing, which cannot be estimated by the traditional camera setup due to the limited field-of-view (FOV), to detect obstacles around a vehicle.
Multiple sets of stereo cameras~\cite{wang2012stereo} or catadioptric lenses~\cite{geyer2003conformal, schonbein2014omnidirectional} can be used, but the size and cost of the multiple cameras or blind spots between $360\degree$ FOV lenses could be problems.
Recently, a wide-baseline omnidirectional stereo setup which consists of four 220$\degree$ FOV fisheye cameras facing the four cardinal directions has been presented~\cite{won2019sweepnet}.
This wide-baseline setup enables long-range sensing while having a full 360$\degree$ coverage, and due to the sufficient overlaps between stereo pairs, most areas are visible from more than two cameras.
Using the same capture system, a robust visual odometry (VO) ROVO~\cite{seok2019rovo}, and an end-to-end deep neural network for the omnidirectional depth estimation OmniMVS~\cite{won2019omnimvs} have been also proposed.

In this paper, we propose an omnidirectional localization and dense mapping system for the wide-baseline omnidirectional stereo setup.
We integrate the omnidirectional dense depth and the pose estimates into a unified dense mapping system while being more efficient and accurate.
We adopt the OmniMVS~\cite{won2019omnimvs} for the omnidirectional depth estimation which estimates a whole $360\degree$ depth map from the input fisheye images.
However, since the network warps unary features extracted from each input image onto 3D global spheres, it requires a huge GPU memory and computational resources to handle the multiple 4D feature volumes.
Therefore, we propose light-weighted and also improved networks using our new training strategy.
We also present improved versions of ROVO~\cite{seok2019rovo} by integrating the depth estimates into the pose estimation process.
Using the omnidirectional dense depth output by the network, we boost the performance of the inter-view matching and triangulation. 
Besides, we implement a loop closing module specialized for our omnidirectional capture system. 
Finally, we fuse the omnidirectional depth and the pose estimation into a truncated signed distance function (TSDF) volume~\cite{curless1996volumetric} to obtain a global 3D map.
Figure~\ref{fig:intro} shows an example 3D map of a challenging indoor environment reconstructed by our proposed system. 

The main contributions are summarized as:
\begin{enumerate}[label=(\roman*)]
\setlength{\topsep}{0pt}
\setlength{\itemsep}{0pt}
\setlength{\partopsep}{0pt} 
\item 
We propose light-weighted and improved networks for the omnidirectional depth estimation from the wide-baseline stereo setup.
The accuracy, the number of parameters, and the running time of our network have become better than the previous versions, which makes our system more practical.
\item
We build a robust omnidirectional visual SLAM system by integrating the depth map into ROVO and adding a loop closing module.
The accuracy of the estimated trajectory is improved than the previous versions in both challenging indoor and large scale outdoor environments.
\item
We present an integrated omnidirectional localization and dense mapping system, and the extensive experiments on synthetic, and real-world indoor and outdoor environments show that our proposed system generates well reconstructed 3D dense maps for various scenes.
\end{enumerate}

\section{RELATED WORK}

Most image-based 3D dense mapping systems follow two steps: dense depth estimation, and temporal fusion of the estimated depths using camera poses.
For the dense depth estimation, temporal or motion stereo methods using a monocular camera can be used for static scenes~\cite{pollefeys2008detailed,hane2015obstacle, schops2017large}, whereas multi-camera systems including stereo cameras are mainly used for more general scenes. 
Recent deep learning-based methods~\cite{kendall2017end,chang2018pyramid,ilg2018occlusions,zhang2019ga} perform stereo depth estimation from the traditional stereo pinhole cameras, which assumes a pair of rectified images as input, in an end-to-end fashion.
Without rectifying or undistorting input images, the plane-sweeping stereo~\cite{collins1996space,gallup2007real,hane2014real} allows for dense stereo matching among multi-view images.
Adopting the plane-sweeping stereo, Cui~\etal~\cite{cui2019real} propose a real-time dense depth estimation from multiple fisheye cameras which have a wider FOV than pinhole cameras.
For the camera pose estimation, several visual odometry methods~\cite{forster2014svo,mur2015orb,cremers2017direct,yokozuka2019vitamin,engel2017direct} have been proposed for monocular systems, however, due to the limitation of monocular setup, the metric scale of the poses cannot be estimated.
Mur-Artal and Tard\'os~\cite{mur2017orb}, and Wang~\etal~\cite{wang2017stereo} propose stereo camera-based systems estimating metric scale poses.
Meanwhile, fisheye cameras are used also for more robust pose estimation.
Caruso~\etal~\cite{caruso2015large} propose a fisheye camera-based visual simultaneous localization and mapping (SLAM) with direct methods which optimize photometric errors of images.
Fisheye stereo camera-based visual odometry systems have been also proposed by Liu~\etal~\cite{liu2017direct} and Matsuki~\etal~\cite{matsuki2018omnidirectional}.

\begin{figure}[t!]
    \centering
    \includegraphics[width=\linewidth]{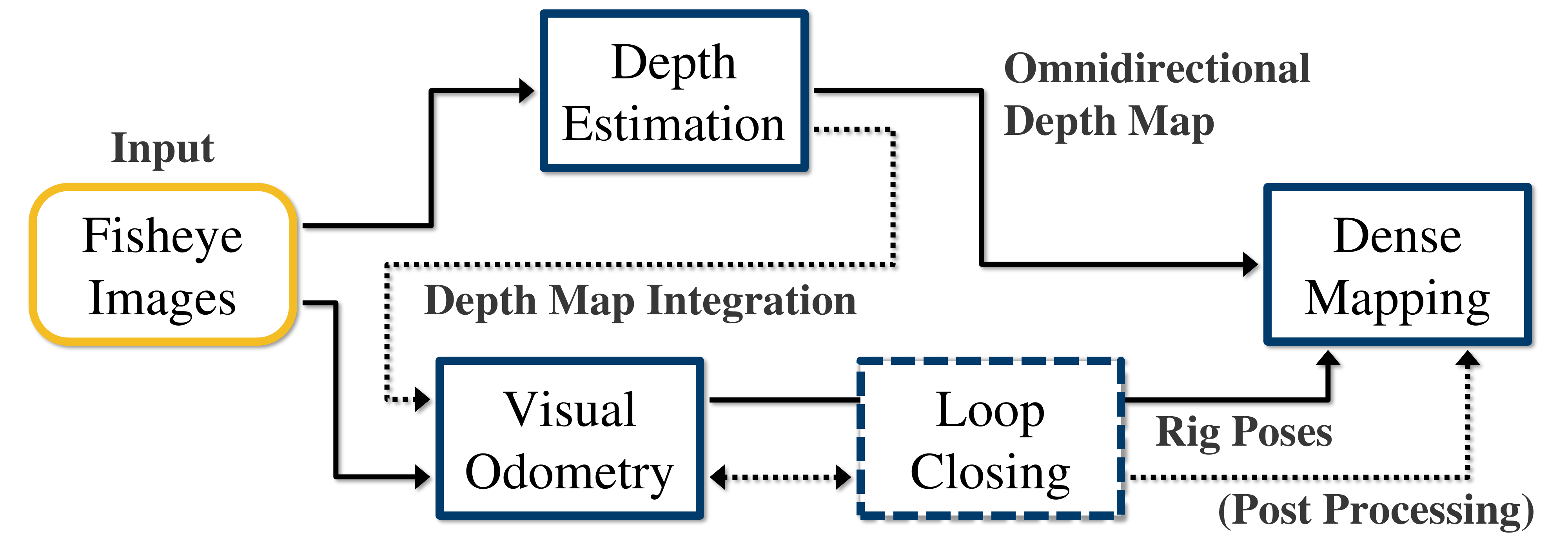}
    \caption{Flowchart of the proposed system. We first estimate omnidirectional depth maps and poses with given fisheye images. The depth map is integrated into the visual odometry, when available. We fuse output depth maps and poses into a TSDF volume to build the 3D map. Corrected poses by the loop closing module are used for building a globally consistent map in the post processing.}
    \label{fig:flowchart}
\vspace{-15pt}
\end{figure}

There have been relatively few systems and works for omnidirectional sensing.
Using 360$\degree$ FOV catadioptric stereo cameras, Geyer~\etal~\cite{geyer2003conformal} and Sch\"onbein~\etal~\cite{schonbein2014omnidirectional} compute stereo disparity from a pair of rectified omnidirectional images.
Pretto~\etal~\cite{pretto2011omnidirectional} propose a triangulation-based omnidirectional dense mapping system using a catadioptric camera and vehicle wheels' motion.
Multi-camera systems have been also proposed for the omnidirectional stereo depth estimation~\cite{wang2012stereo} and the visual odometry~\cite{liu2018towards}, however, they require many cameras which could be problematic in some cases.
Recently using only a few fisheye cameras, a wide-baseline omnidirectional stereo system has been proposed~\cite{won2019sweepnet}.
Using the same rig setup, Won~\etal~\cite{won2019omnimvs} propose an end-to-end network for the omnidirectional depth estimation, and Seok and Lim~\cite{seok2019rovo} present a robust visual odometry.

\section{METHOD}

In this section, we describe our omnidirectional localization and dense mapping system.
Figure~\ref{fig:flowchart} illustrates the overall procedure of our proposed system consisting of three main modules: omnidirectional depth estimation, omnidirectional visual SLAM, and TSDF-based dense mapping.
%

\subsection{Omnidirectional Depth Estimation}
\label{subsec:omnidepth}
We use the wide-baseline multi-camera rig system and the spherical sweeping-based approach~\cite{won2019sweepnet} for the omnidirectional stereo matching.
In such a wide-baseline setup, spherical sweeping stereo finds stereo correspondences from the multi-view images by warping them onto the global sweeping spheres centered at the rig origin, and the radius of each sweeping sphere corresponds to an inverse depth hypothesis for each ray $\mb{p}(\theta,\phi)=(\cos{(\phi)}\cos{(\theta)},\sin{(\phi)},\cos{(\phi)}\sin{(\theta)})^{\top}$ in the warped $W\times H$ equirectangular images where $(\theta, \phi)$ is the spherical coordinate of $\mb{p}$.

Recently, an end-to-end network OmniMVS~\cite{won2019omnimvs} which consists of unary feature extraction, spherical sweeping, and cost volume computation modules, allows for dense omnidirectional multi-view stereo matching from the wide-baseline rig.
To estimate an omnidirectional depth, $\frac{1}{2}W_I\times \frac{1}{2}H_I\times C$ unary feature maps are extracted from the input fisheye images using the residual blocks~\cite{he2016deep} where $W_I$ and $H_I$ are the size of the images, and $C$ is the number of channels.
The feature maps then are warped onto the global spheres using $4\times(W\times H\times (N/2))$ sweeping $x,y$ coordinates where $N$ is the number of inverse depths, and are concatenated to build the 4D omnidirectional matching cost volume.
The 3D encoder-decoder architecture~\cite{kendall2017end} computes and regularizes the cost volume.
Finally, one minimum index of inverse depths for each ray is picked by the softargmin operation while considering the multiple true matches which may occur in such a global sweeping approach~\cite{won2019omnimvs}.

\subsec{Light-weighted OmniMVS}
We use the OmniMVS~\cite{won2019omnimvs} to acquire a dense omnidirectional depth.
However the size of tensor in the network grows up to $\frac{1}{2}W\times \frac{1}{2}H\times \frac{1}{2}N\times 4C$ ($C=32$ in~\cite{won2019omnimvs}) which requires a huge GPU memory and computational resources.
To develop light-weight networks, we reduce the number of channels $C$ of all layers several times, and to train the network more effectively with the small number of channels, we guide the network by adding a 2D convolutional layer without ReLU into the end of the feature extraction module.
This enables the networks to discriminate the negative features and the invisible area in the warped spherical feature maps, which are set to 0 by ReLU and warping process respectively.
We also apply the disparity regression~\cite{kendall2017end} to regress inverse-depth indices by the weighted summation of all the candidates rather than picking one minimum index.
In this way, the discretization errors can be reduced as shown in Fig.~\ref{fig:qual_depth}.
In order to train the network, we use the smooth L1 loss as
\[
\mathcal{L}(\mb{p})=\frac{1}{M}\sum{L_{1;smooth}(n(\mb{p})-n^*(\mb{p}))},
\]
where $n$ and $n^*$ are the estimated and ground-truth inverse-depth index respectively, and $M$ is the number of valid pixels.
We use the stochastic gradient descent with momentum to minimize the loss.

\subsection{Ominidrectional Visual SLAM}
\label{subsec:rovo}
Localization is also an essential part of the 3D dense mapping and recently proposed ROVO~\cite{seok2019rovo} robustly estimates rig poses of the wide-baseline omnidirectional stereo system.
In~\cite{seok2019rovo}, ROVO follows four steps: hybrid projection; intra-view tracking, and inter-view matching; pose estimation; and joint optimization.
At first, input fisheye images are warped onto the hybrid projection images in which the ORB features~\cite{rublee2011orb} are detected.
Secondly, the detected ORB features are temporally tracked using KLT~\cite{lucas1981iterative}, and the inter-view feature matching between adjacent cameras is performed.
Both tracked and matched features are then triangulated to each corresponding 3D point.
Thirdly using the 2D-3D feature correspondences, the rig pose is initialized by multi-view P3P RANSAC~\cite{seok2019rovo} and optimized by pose-only bundle adjustment (BA)~\cite{triggs1999bundle}.
Finally, the estimated poses and observed 3D points in the local window are simultaneously optimized by local bundle adjustment (LBA).

\subsec{Depth Map Integration}
\label{subsec:rovo-depth}
We adopt ROVO~\cite{seok2019rovo} as a baseline of our localization module, however, triangulated 3D points in the feature tracking stage may have large uncertainty since it is highly dependent on the camera motion.
Moreover, the inter-view matching process, which finds correspondences from unspecified areas in the other view, is inefficient and easy to mismatch them.
Therefore, we integrate the omnidirectional depth map estimated in Section~\ref{subsec:omnidepth} into our localization module ROVO$^+$ for better efficiency and robustness.
We first replace the triangulated depth with the network's estimation to avoid the uncertainty.
Second, in the inter-view feature matching stage, we reproject the features to other adjacent views using the depth and find the best match around the projected location by using the nearest neighbor search.
These processes give better initialization of depth and help to avoid local minimum in the joint optimization process: pose-only BA and LBA.
Considering the running time of the depth estimation as described in Table~\ref{tab:omnidepth}, both integration processes are activated only in the depth-available frames.
Due to the high-accuracy of the depth estimation, these integration processes improve the overall performance of the visual odometry.


%
\subsec{Loop Closing}
\label{subsec:rovo-loop}
Loop closing has been widely used in many SLAM systems to correct large drifts of estimated trajectory.
We propose a loop closing module for the wide-baseline omnidirectional stereo system while utilizing the benefits of the wide FOV.
First, we use the vocabulary tree ~\cite{GalvezTRO12} trained in~\cite{mur2015orb,mur2017orb} for detecting loop closures.
We create the query for the vocabulary tree by stacking all feature descriptors observed by each camera to take full advantages of the 360$\degree$ FOV of the rig system in place recognition.
%
%
After selecting the loop candidates, we check the geometric consistency of them by using the multi-view P3P RANSAC~\cite{seok2019rovo}.
When matching features between current and candidate's views, we only consider the circular shift of the rig (e.g., 1-2-3-4, 2-3-4-1, 3-4-1-2, or 4-1-2-3), assuming the rig does not flip or skew.
Finally, we process the pose-graph optimization to correct the trajectory.
Since a local 3D map is enough for path planning and obstacle detection~\cite{cui2019real}, we use the corrected trajectory for building a global map in post-processing.

\subsection{TSDF-based Dense Mapping}
In order to obtain a global 3D map, we fuse the omnidirectional depth maps and the rig poses estimated in Section~\ref{subsec:omnidepth} and~\ref{subsec:rovo} into the truncated signed distance function (TSDF) volume~\cite{curless1996volumetric}.
The volume consists of a set of voxels containing a truncated signed distance to an object surface.
We reproduce the TSDF integrator in Voxblox library~\cite{oleynikova2017voxblox} which runs in real-time on a single CPU core.

When a new omnidirectional depth arrives, we convert it to a 3D pointcloud and transform it into the global coordinate system using the estimated rig position.
We then cast a ray from the estimated current rig position to every point in the pointcloud, and update the voxels along the ray so as to remove noisy surfaces or moving objects by voxel carving.
We update the voxel's existing distance and weight values, $D$ and $\Gamma$ with a new observed distance $d(\mb{X},\mb{P},\mb{O_r})=|\mb{P}-\mb{O_r}|-(\mb{P}-\mb{O_r})\bullet(\mb{X}-\mb{O_r})$, 
where $\mb{P}$ is the position of a 3D point, $\mb{O_r}$ is the rig position, and $\mb{X}$ is the center position of the voxel.
We also use the linear drop-off strategy~\cite{bylow2013real,oleynikova2017voxblox} for updating the weight $\gamma$ as
\[
\gamma =
    \begin{cases}
    \rho & -v < d \\
    \frac{\rho}{\delta-v}(d+\delta) & -\delta < d < -v \\
    0 & \phantom{-v < } d < -\delta
    \end{cases},
\]
where $\rho$ is the initial weight parameter, $v$ is the voxel size, and $\delta$ is the truncation distance.
Using new observations, the new distance and weight are updated as
\begin{align}
D(\mb{X}) ~&\leftarrow~  \frac{\Gamma(\mb{X})D(\mb{X})+\gamma(\mb{X},\mb{P},\mb{O_r})d(\mb{X},\mb{P},\mb{O_r})}{\Gamma(\mb{X})+\gamma(\mb{X},\mb{P},\mb{O_r})} \\
\Gamma(\mb{X}) ~& \leftarrow~ \Gamma(\mb{X}) + \gamma(\mb{X},\mb{P},\mb{O_r})
\end{align}

\begin{table}[tb!]
\resizebox{\linewidth}{!}{%
\begin{tabular}{lrrrrrr} \bottomrule
\multirow{2}{*}{Network} & \multicolumn{1}{c}{\multirow{2}{*}{Sunny}} & \multicolumn{1}{c}{\multirow{2}{*}{Cloudy}} & \multicolumn{1}{c}{\multirow{2}{*}{Sunset}} & \multicolumn{1}{c}{Omni} & \multicolumn{1}{c}{\# Param.} & \multicolumn{1}{c}{Run} \\
 & \multicolumn{1}{c}{} & \multicolumn{1}{c}{} & \multicolumn{1}{c}{} & \multicolumn{1}{c}{House} & \multicolumn{1}{c}{($\si{M}$)} & \multicolumn{1}{c}{Time ($\si{s}$)} \\ \hline \hline
OmniMVS~\cite{won2019omnimvs} & 0.79 & 0.72 & 0.79 & 1.04 &  10.91 & 0.66 \\
OmniMVS$^+$ & \tb{0.31} & \tb{0.30} & \tb{0.32} & \tb{0.61} &  10.91 & 0.66 \\
Small$^+$ & 0.48 & 0.45 & 0.50 & 0.69 & 0.68 & 0.21 \\
Tiny$^+$ & 0.64 & 0.61 & 0.66 & 0.94 &  \tb{0.17} & \tb{0.13} \\ \toprule
\end{tabular}
}
\caption{Quantitative evaluation of the networks. We use the mean absolute error of the inverse depth index. The errors are averaged over all test frames.}
\label{tab:omnidepth}
\vspace{-5pt}
\end{table}

\begin{figure}[t]
    \includegraphics[width=\linewidth]{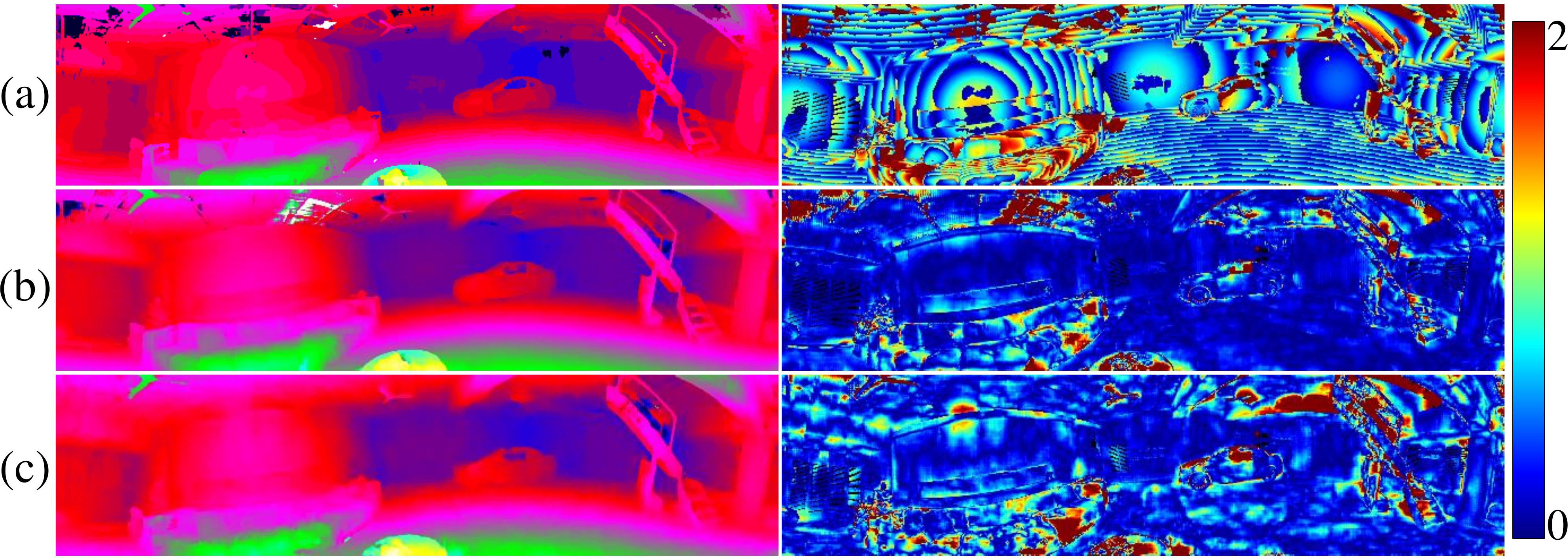}
    \caption{Comparison of the networks on the Garage, (a) OmniMVS~\cite{won2019omnimvs}, (b) OmniMVS$^+$, and (c) Tiny$^+$. From left: estimated inverse depth index, and error map of (\ref{eq:omni_error}) ranges from 0 to 2.} 
    \label{fig:qual_depth}
 \vspace{-15pt}   
\end{figure}

We also add a parameter representing the number of observations into each voxel to consider unreliable estimates on the hard regions, for example, textureless or differently appeared surfaces due to the wide-baseline of the capture system.
Meanwhile, reliable areas can be sufficiently observed due to the 360$\degree$ FOV of the depth sensing.
We use the marching cubes surface construction~\cite{lorensen1987marching} for the visualization.

\section{EXPERIMENTAL RESULTS}

\subsection{Experimental Setup}


We evaluate our proposed methods on the real data: IT/BT and Wangsimni, and the synthetic datasets: Sunny, Cloudy, Sunset, and Garage.
For the real data, we use four cameras with $220\degree$ fisheye lenses as~\cite{won2019sweepnet}.
The capture system captures $4\times(1600\times1532)$ gray images synchronized by the software trigger at up to 20 Hz.
We calibrate the intrinsic and extrinsic parameters of the rig using a checkerboard~\cite{scaramuzza2006flexible,urban2015improved,won2019sweepnet}.
\begin{table}[ht!]
\resizebox{\linewidth}{!}{%
\begin{tabular}{lcclr}\bottomrule
\multirow{2}{*}{Dataset} & \multirow{2}{*}{ROVO~\cite{seok2019rovo}} & \multirow{2}{*}{ROVO$^+$}  & \multicolumn{1}{c}{\multirow{2}{*}{Metric}} & \multicolumn{1}{c}{Total} \\
 & & & & \multicolumn{1}{c}{Length} \\
\hline \hline
Sunny                    & 0.54              & \tb{0.20}                 & \multirow{4}{*}{$\left.\vphantom{\begin{tabular}{c}.\\.\\.\\.\end{tabular}}\!\! \right\}$\begin{tabular}[c]{@{}l@{}}~ATE$_{trans}$~\\~RMSE~($\si{m}_s$)\end{tabular}}        & \multirow{3}{*}{$\left.\vphantom{\begin{tabular}{c}.\\.\\.\end{tabular}}\!\! \right\}~\SI{426}{m}_s$}   \\
Cloudy                   & 0.81              & \tb{0.28}               &     &   \\
Sunset                   & 0.78              & \tb{0.29}              &     &   \\
Garage                   & 0.0011            & \tb{0.0010}             &     & $\SI{28}{m}_s$    \\ 
IT/BT                    & 1.28              & \tb{0.78}              & Start-to-End $(\si{m})$   & $\SI{230}{m}_{~}$      \\ \toprule
\end{tabular}
}
\caption{Quantitative evaluation of the visual SLAM. In all datasets, ROVO$^+$ shows better performance than ROVO~\cite{seok2019rovo}.}
\label{tab:omnivo}
\vspace{-15pt}
\end{table}
IT/BT sequence is captured by a person walking around a duplex building and returning to the starting position while holding a small square-shaped rig~($0.3\times\SI{0.3}{m}$) and has a length of about $\SI{230}{m}$.
Wangsimni sequence is captured by the outdoor rig installed on a minivan moving around the very narrow alley and has a length of about $\SI{2.6}{\km}$, and we mask out the minivan area from the input fisheye images while sphere sweeping and adapting hybrid projection.
For the synthetic data, realistically rendered $800\times768$ input images are converted to grayscale, and we define scaled meter $\si{m}_s$ by referring 3D models such as cars since the synthetically rendered data have no actual unit.
Sunny, Cloudy, Sunset datasets are realistic cityscape datasets proposed in~\cite{won2019omnimvs}.
They have different photometric environments and a length of $\SI{426}{m}_s$.
Garage is a realistic indoor dataset which is very short ($\SI{28}{m}_s$) and has large textureless regions.
In the Sunny, Cloudy, and Sunset, $\SI{1}{m}_s = 100/3$, and $\SI{1}{m}_s = 100$ in the Garage.
We also use the ground-truth intrinsic and extrinsic parameters of the virtual rig in the synthetic datasets.

\begin{figure*}[hbt!]
    \centering
    \resizebox{\linewidth}{!}{%
    \begin{subfigure}[b]{0.245\linewidth}
        \captionsetup{justification=centering}
        \includegraphics[width=\linewidth]{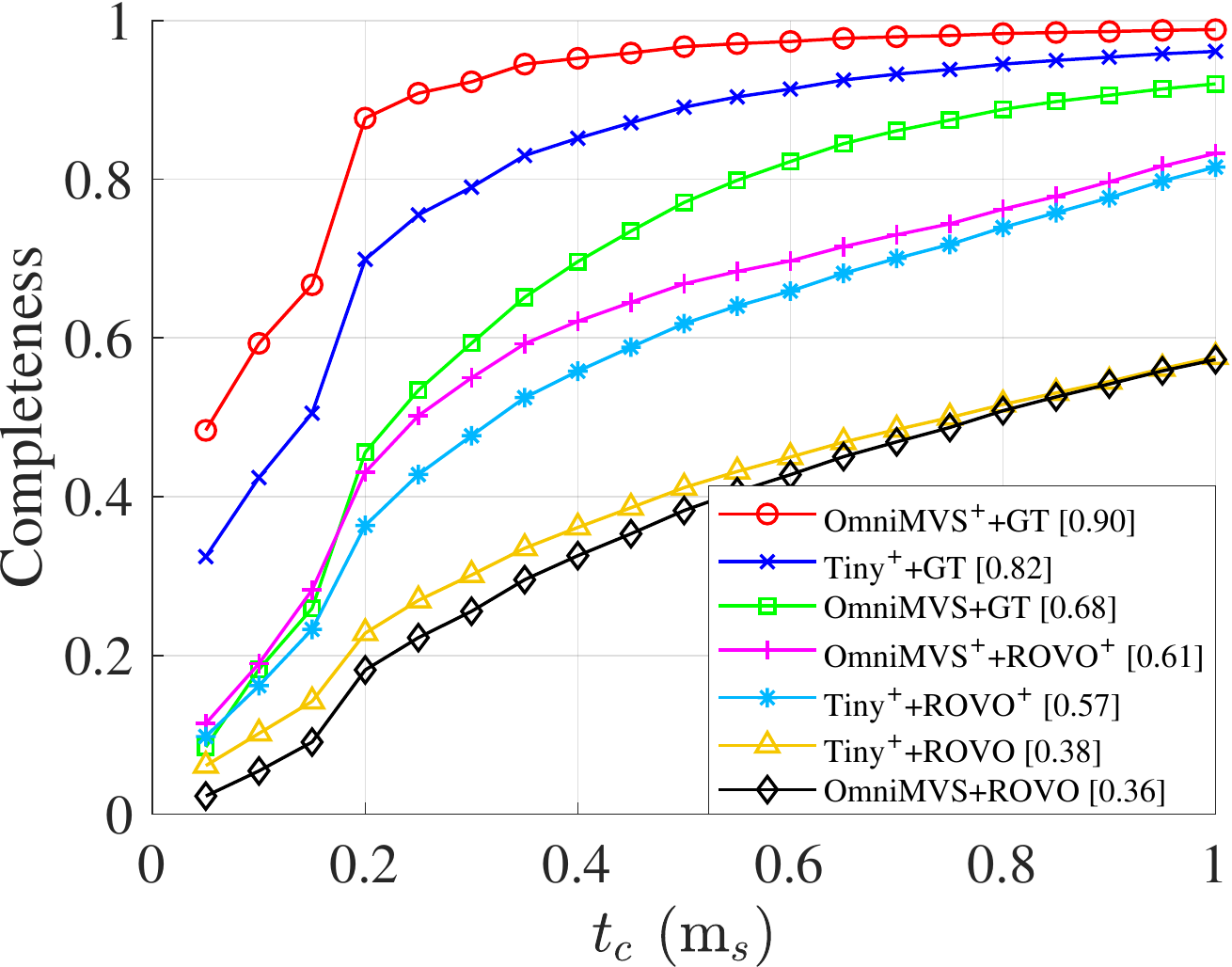}
    \end{subfigure} \hspace{-5pt}
    \begin{subfigure}[b]{0.245\linewidth}
        \captionsetup{justification=centering}
        \includegraphics[width=\linewidth]{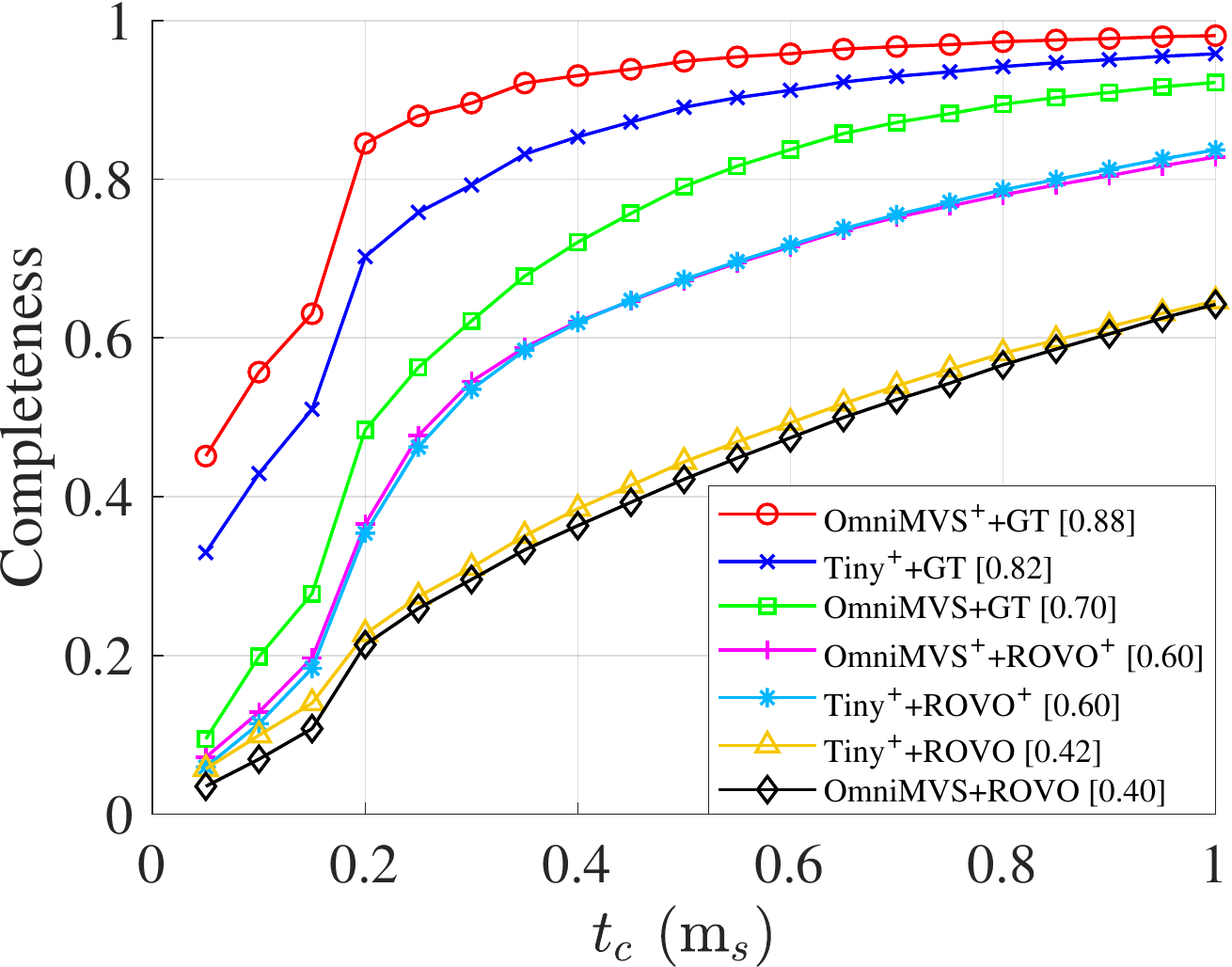}
    \end{subfigure} \hspace{-5pt}
    \begin{subfigure}[b]{0.245\linewidth}
        \captionsetup{justification=centering}
        \includegraphics[width=\linewidth]{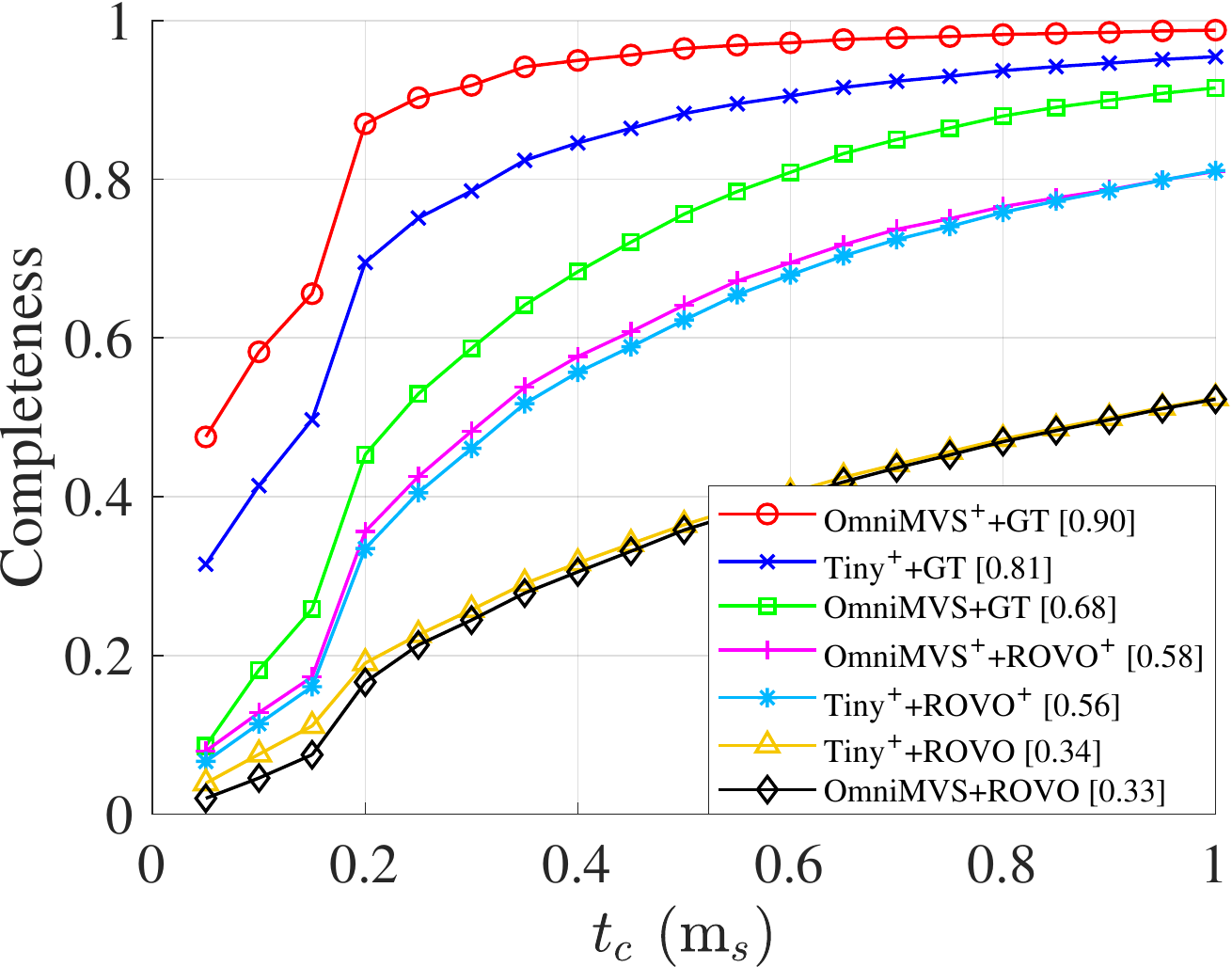}
    \end{subfigure} \hspace{-5pt}
    \begin{subfigure}[b]{0.245\linewidth}
        \captionsetup{justification=centering}
        \includegraphics[width=\linewidth]{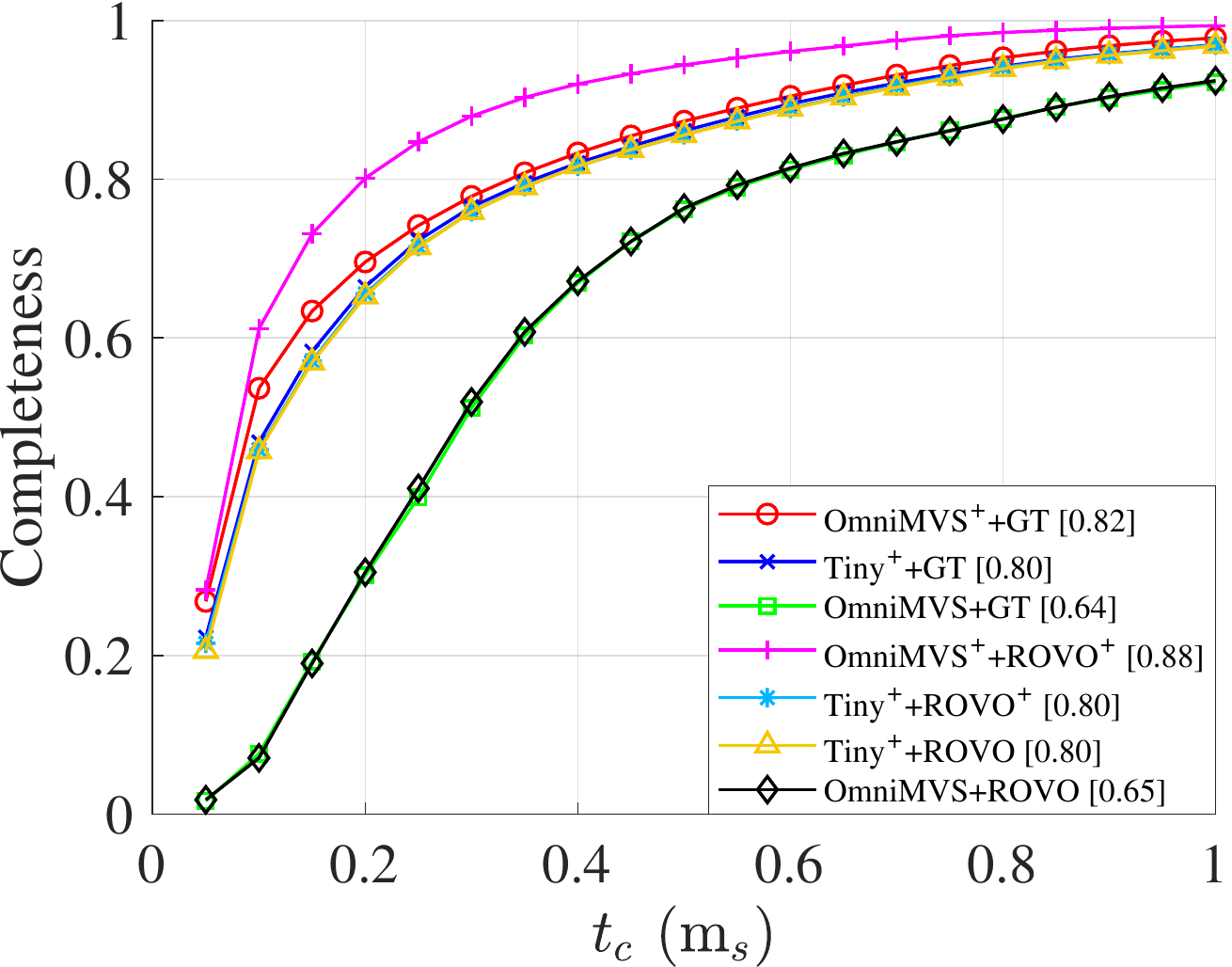}
    \end{subfigure}
    }
    
    \resizebox{\linewidth}{!}{%
    \begin{subfigure}[b]{0.245\linewidth}
        \captionsetup{justification=centering}
        \includegraphics[width=\linewidth]{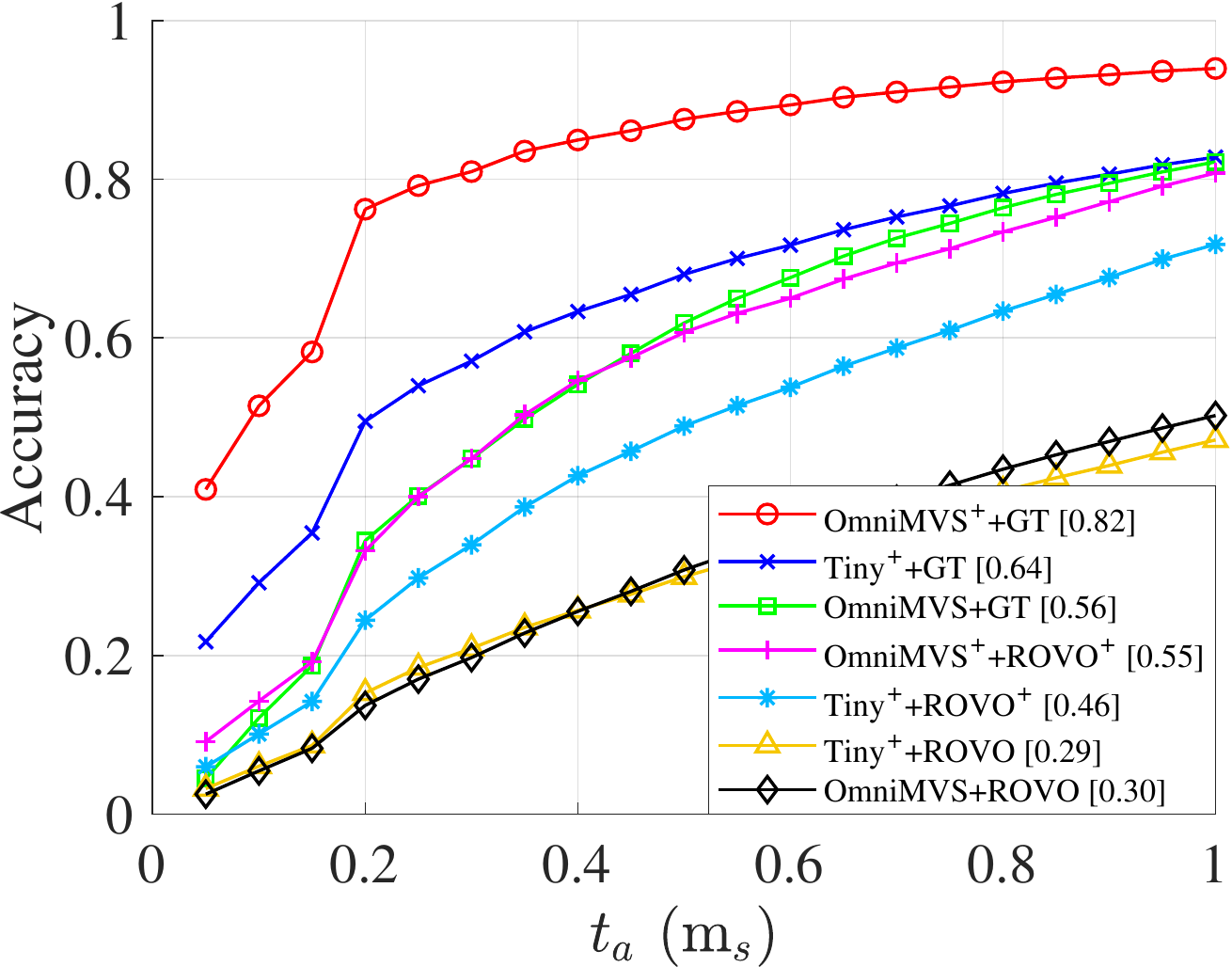}
        \caption{Sunny}\label{fig:sunny_mapping}
    \end{subfigure} \hspace{-5pt}
    \begin{subfigure}[b]{0.245\linewidth}
        \captionsetup{justification=centering}
        \includegraphics[width=\linewidth]{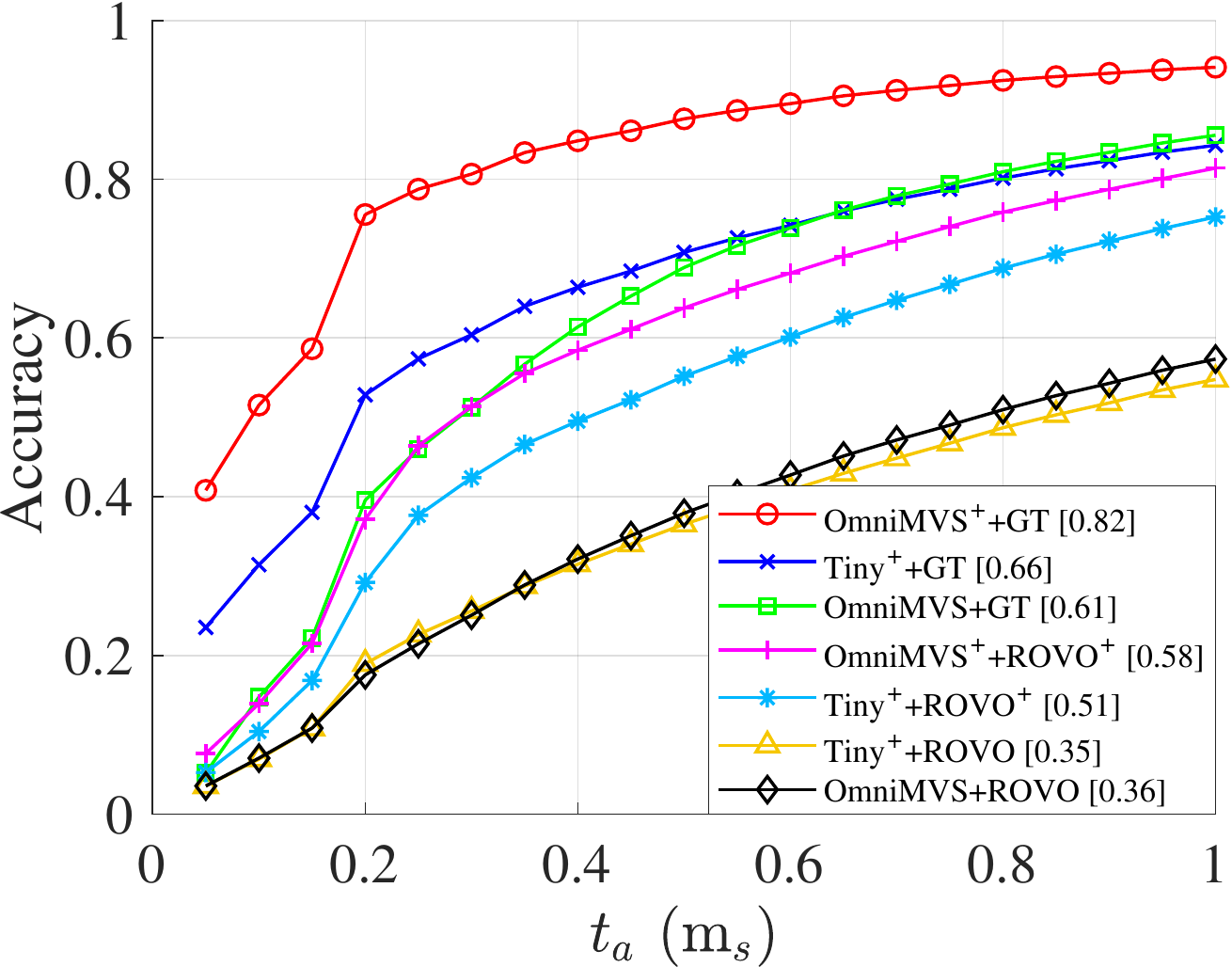}
        \caption{Cloudy}\label{fig:cloudy_mapping}
    \end{subfigure} \hspace{-5pt}
    \begin{subfigure}[b]{0.245\linewidth}
        \captionsetup{justification=centering}
        \includegraphics[width=\linewidth]{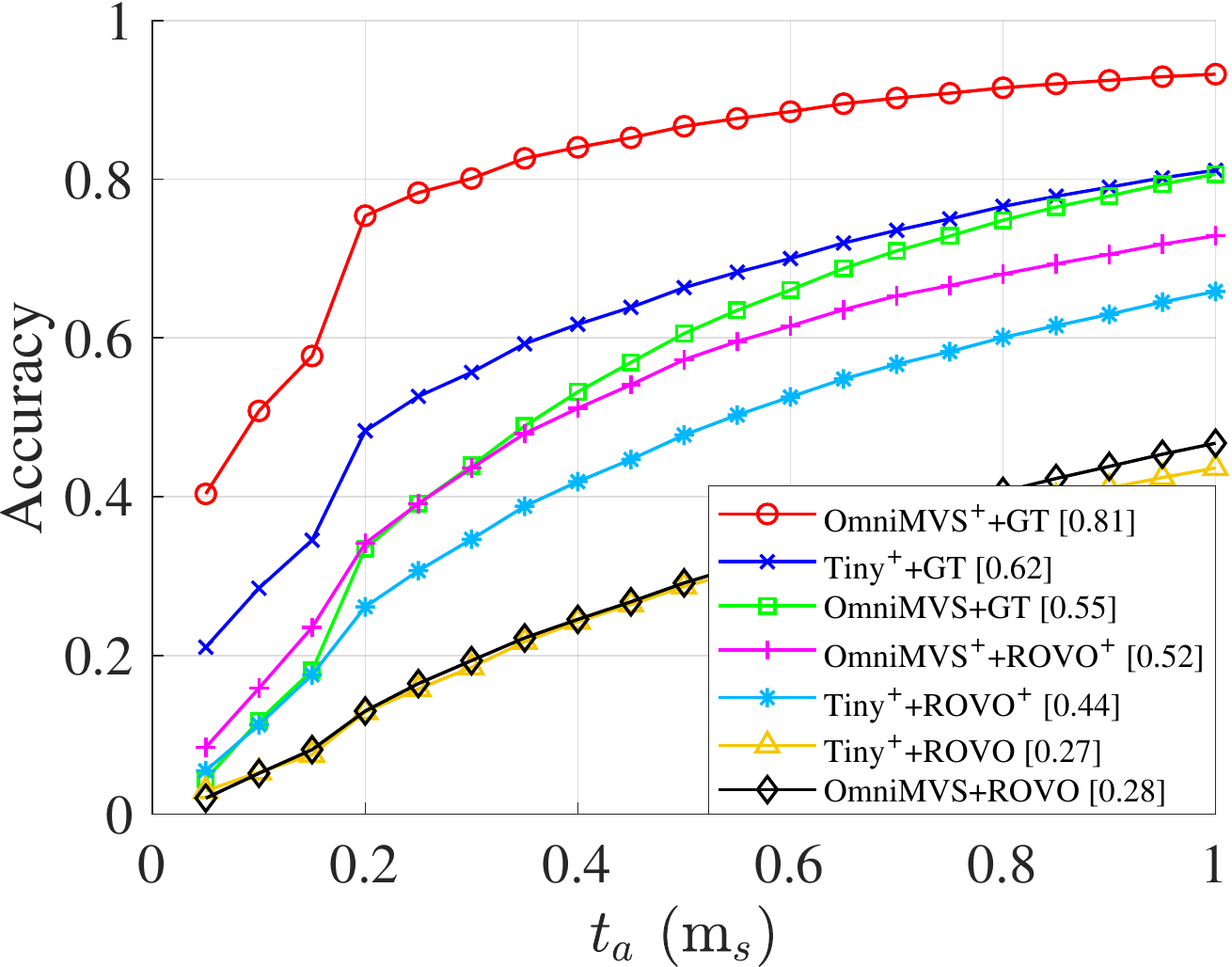}
        \caption{Sunset}\label{fig:sunset_mapping}
    \end{subfigure} \hspace{-5pt}
    \begin{subfigure}[b]{0.245\linewidth}
        \captionsetup{justification=centering}
        \includegraphics[width=\linewidth]{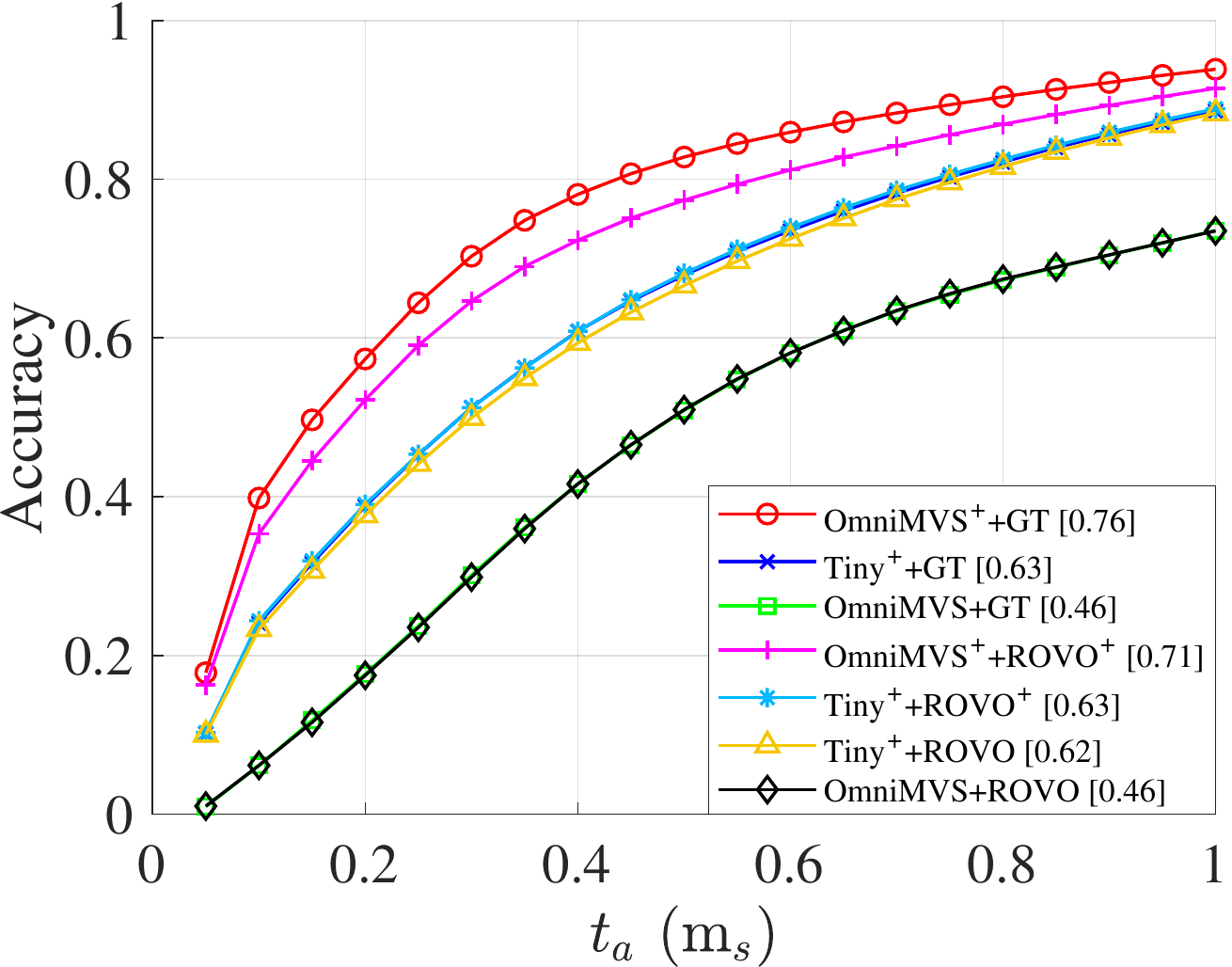}
        \caption{Garage}\label{fig:garage_mapping}
    \end{subfigure}}
    \caption{Evaluation of dense mapping results. Top: completeness, Bottom: accuracy. The mean ratio for each method is shown in the legend. We use OmniMVS~\cite{won2019omnimvs}, OmniMVS$^+$, and Tiny$^+$ for the depth; and the GT trajectory, ROVO~\cite{seok2019rovo}, and ROVO$^+$ for the pose.} 
    \label{fig:eval_mapping}
\vspace{-10pt}
\end{figure*}

\begin{figure}[hbt!]
\resizebox{\linewidth}{!}{%
\includegraphics[width=\linewidth]{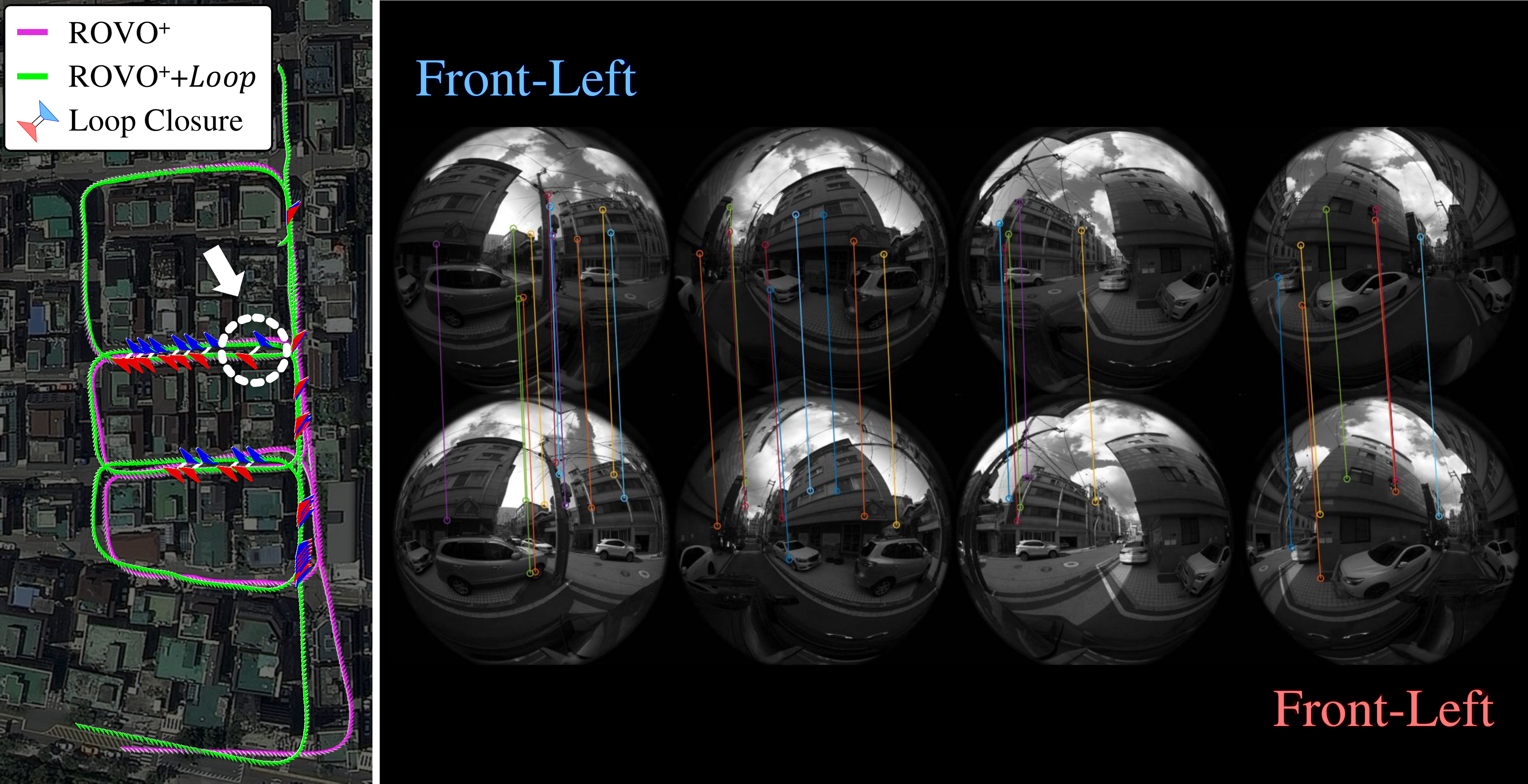}
}
\caption{Effectiveness of our loop closing on the Wangsimni.
Left: comparative results of the trajectories with detected loop closures.
Right: feature inliers of matched keyframes in the loop closure marked as a white circle.
Even the direction of current keyframe is opposite to the matched keyframe, the loop is correctly closed.
}
\label{fig:qual_loop_wangsimni}
\vspace{-15pt}
\end{figure}

\subsection{Evaluation of Omnidirectional Depth Estimation}
We evaluate our proposed networks on the synthetic datasets as~\cite{won2019omnimvs} with the ground-truth depths.
As a baseline network, we use the fine-tuned version of the OmniMVS in~\cite{won2019omnimvs} whose the number of base channels $C$ is set to $32$, and we set the number of base channels $C$ of our networks, OmniMVS$^+$, Small$^+$, and Tiny$^+$ to 32, 8, and 4 respectively. 
We use the percent error of estimated inverse depth index from ground-truth as
\begin{equation}
\label{eq:omni_error}
\mathcal{E}(\mb{p})=\frac{100}{N}|n(\mb{p})-n^*(\mb{p})|.
\end{equation}
We set the size of omnidirectional depth output by the networks to $H=160$ and $W=640$, the number of inverse depth $N=192$, and $\phi$ from $-45\degree$ to $45\degree$ for the IT/BT, Sunny, Cloudy, Sunset, and Garage.
For the Wangsimni, we set the size of omnidirectional depth to $H=128$ and $W=768$ and $\phi$ from $-30\degree$ to $30\degree$.
We also measure running time of each networks on the IT/BT dataset using Nvidia Quadro GV100 GPU.
Table~\ref{tab:omnidepth} shows the mean absolute error of (\ref{eq:omni_error}), the number of parameters of each networks, and the running time.
As shown in Table~\ref{tab:omnidepth}, the networks trained using our strategy significantly improve the performances, and our Tiny$^+$ is $5\times$ faster and much lighter than OmniMVS~\cite{won2019omnimvs} and performs better which makes our system more practical.
From the error maps in Fig.~\ref{fig:qual_depth}, we can clearly observe the discretization error of OmniMVS~\cite{won2019omnimvs}, which is greatly reduced by our proposed networks.

\subsection{Evaluation of Estimated Poses}
 \label{subsec:eval-poses}

In this section, we evaluate the accuracy and robustness of our pose estimation module on both synthetic and real-world datasets.
%
%
To show the improvement of ours against the previous work, we compare our method to~\cite{seok2019rovo}. 
For the experiments, we use the number of features to $4\times600$ in Sunny, Cloudy, Sunset, and $4\times400$ in Garage and IT/BT. 
Note that the number of features is heuristically chosen considering the environmental characteristic, for example, unlike the texture-rich outdoor (Sunny, Cloudy and Sunset), indoor scenes (Garage, IT/BT) mainly consist of large textureless environments such as walls and floors, so that we use a smaller number of feature points for the indoor datasets.
We use two types of the error metric for the quantitative evaluation: the absolute trajectory error of translation (ATE$_{trans}$) and the Start-to-End error.
ATE$_{trans}$ is a widely used metric for measuring accuracy of VO/SLAM algorithms and is calculated by the root mean squared error (RMSE) after aligning the estimated poses to the ground-truth poses.
Start-to-End error is measured by the difference between the start and end position.
%
%
%
%
%
As shown in Table~\ref{tab:omnivo}, our method shows better accuracy than that of ROVO in overall.
In Sunny, Sunset, Cloudy datasets, the errors of ROVO$^+$ are reduced by $0.4\times$ that of ROVO on average.
In Garage dataset, ROVO$^+$ shows a slightly better result than that of ROVO.
These results show that using the depth information has benefits of improving the VO performance.
%

\subsec{Effectiveness of Loop Closing}
To prove the effectiveness of our loop closing module, we use a part of the Wangsimni dataset.
We show the qualitative comparison of the estimated rig poses with and without the loop closing in Fig.~\ref{fig:qual_loop_wangsimni}.
While the rotational drifts exist in ROVO$^+$, the trajectory of ROVO$^++Loop$ is well aligned to the satellite map.
%
%

\begin{figure*}[hbt!]
    \centering
    \resizebox{\linewidth}{!}{%
    \begin{subfigure}[b]{0.69375\linewidth}
        \captionsetup{justification=centering}
        \includegraphics[width=\linewidth]{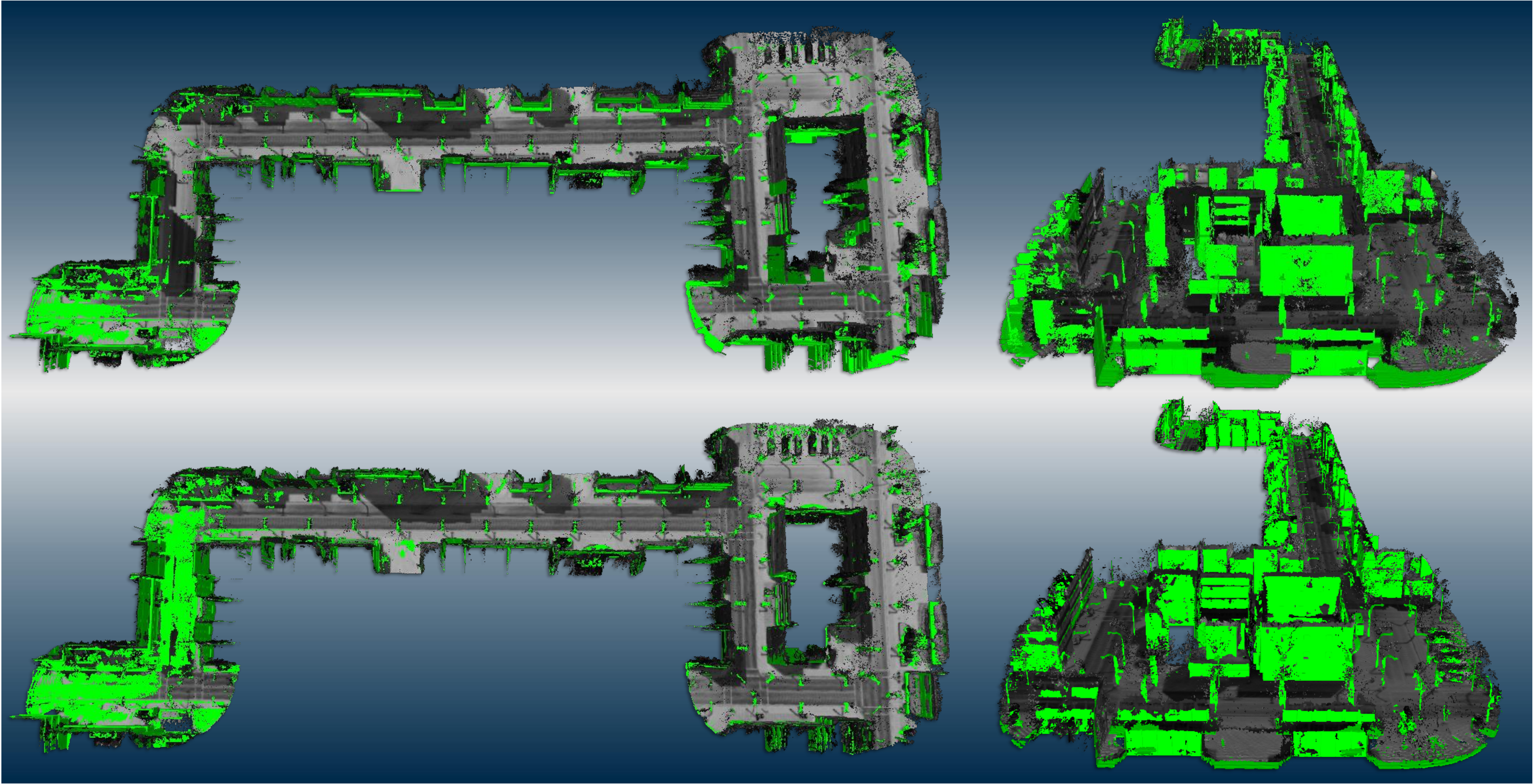}
        \caption{Tiny$^+$+ROVO~\cite{seok2019rovo} (top) and Tiny$^+$+ROVO$^+$}\label{fig:qual_mapping_sunny}
    \end{subfigure}
    \begin{subfigure}[b]{0.296875\linewidth}
        \captionsetup{justification=centering}
        \includegraphics[width=\linewidth]{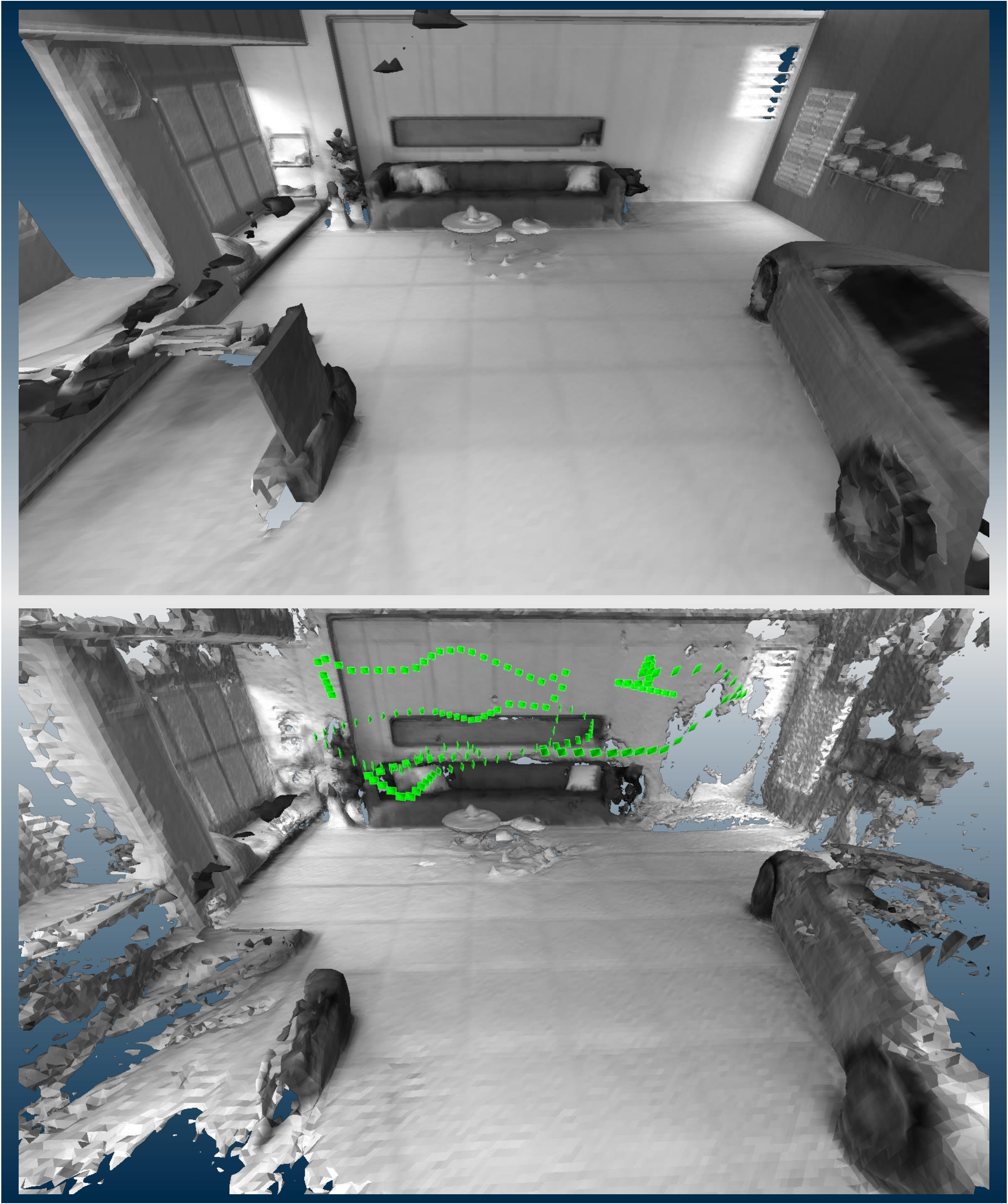}
        \caption{GT (top) and OmniMVS$^+$+ROVO$^+$}\label{fig:qual_mapping_garage}
    \end{subfigure}
    }
    \caption{Qualitative dense mapping results on the synthetic datasets. (a) Results on the Sunny. Green represents GT map. Our ROVO$^+$ reduces the drift errors of the estimated poses. (b) Results on the Garage. Green represents the estimated trajectory of the front camera.} 
    \label{fig:qual_mapping}
    \vspace{-5pt}
\end{figure*}

\begin{figure*}[hbt!]
    \centering
    \includegraphics[width=\linewidth]{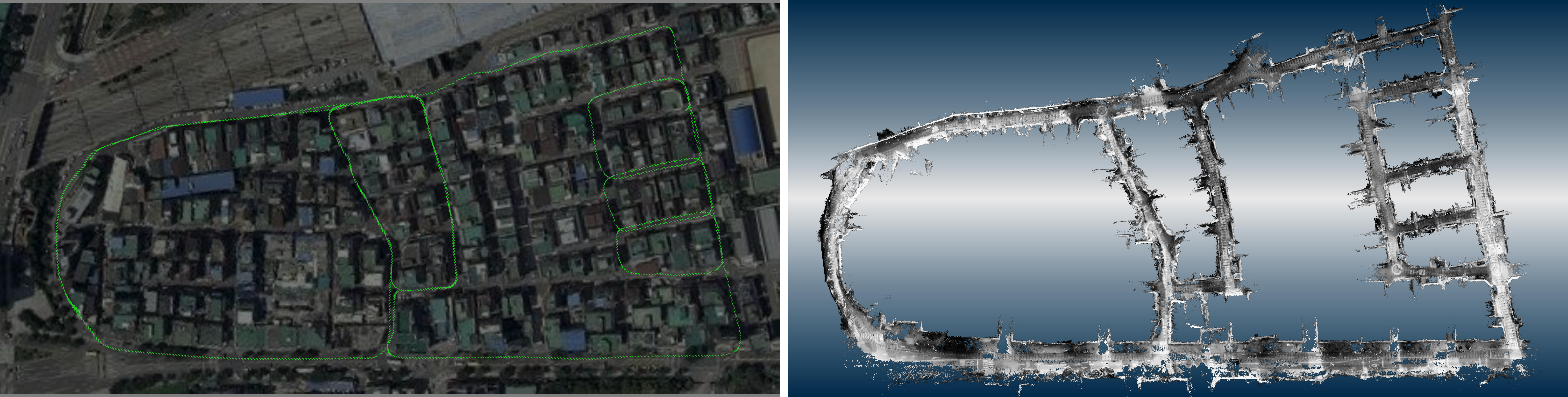}
    \caption{Qualitative results on the Wangsimni dataset. Left: estimated trajectory orthographically projected on the satellite image. Right: corresponding dense mapping result. We apply the histogram equalization to the vertices' color for the visualization.}
    \label{fig:qual_dense_wangsimni}
    \vspace{-15pt}
\end{figure*}

\subsection{Evaluation of Dense Mapping}
In order to evaluate the 3D map output by our system, we reconstruct the ground-truth 3D maps by registering the ground-truth omnidirectional depth maps and rig poses of all the frames into the TSDF voxel grid.
We also render semantic level segmentation masks and remove dynamic objects such as moving cars from the ground-truth depth maps using the masks.
We use completeness and accuracy~\cite{schops2017multi, cui2019real} for the evaluation criteria and evaluate the methods on the Sunny, Cloudy, Sunset, and Garage datasets.

Given two sets of reconstructed mesh's vertices $V$ and $V'$, the distance of vertex $\mb{v} \in V$ to its closets vertex $\mb{v'} \in V'$ is defined as $\mathcal{D}_{\mb{v}}(V,V')$.
Then the completeness is defined as 
$\mathcal{C}_{t_c}(V,V^*)=|\{\mb{v^*}~|~ \mathcal{D}_{\mb{v^*}}(V^*,V) < t_c \}|/|V^*|$,
and the accuracy is defined as
$\mathcal{A}_{t_a}(V,V^*)=|\{\mb{v}~|~  \mathcal{D}_{\mb{v}}(V,V^*) < t_a \}|/|V|$,
where $V$ and $V^*$ are the estimated and ground-truth 3D maps respectively, $t_c$ and $t_a$ are the distance thresholds, and $|\cdot|$ represents the number of points in a set.
%
We reconstruct the 3D map by using the omnidirectional depth from OmniMVS~\cite{won2019omnimvs}, our OmniMVS$^+$, and Tiny$^+$. We use ground-truth poses, ROVO~\cite{seok2019rovo}, and our ROVO$^+$ for the pose without loop closing, and only the selected keyframes by the odometers are used for the dense mapping.
The TSDF voxel size $v$ is set to $\SI{0.15}{m}_s$, and the weight parameter $\rho$ is set to $1$ for Sunny, Cloudy, and Sunset, and $\SI{0.05}{m}_s$ for Garage, and for the Garage, we only use voxels whose number of observations is more than 3 since it has large textureless regions.

Figure~\ref{fig:eval_mapping} shows the completeness and the accuracy graph of the dense mapping results on the synthetic datasets.
Our networks OmniMVS$^+$ and Tiny$^+$ reconstructs the 3D map more precisely than the OmniMVS~\cite{won2019omnimvs} especially for the lower thresholds as shown in Fig.~\ref{fig:qual_depth}.
Using poses from ROVO$^+$ also performs better than ROVO~\cite{seok2019rovo} in all datasets, and Figure~\ref{fig:qual_mapping_sunny} shows the qualitative comparison between them.
The qualitative result on the Garage is also shown in Fig.~\ref{fig:qual_mapping_garage}, and our method successfully reconstructs the small objects such as table and sofa, and also the large textureless regions like walls and the floor.
We also show qualitative results of our system on the real-world datasets: IT/BT and Wangsimni, in Fig.~\ref{fig:intro} and~\ref{fig:qual_dense_wangsimni} respectively.
We use the depths from OmniMVS$^+$ and the poses from ROVO$^{+}+Loop$, and only use voxels whose number of observations is more than 3 for both datasets.
For Wangsimni, we reduce the resolution of input images in half and apply gamma correction to them.
The TSDF voxel size $v$ is set to $\SI{0.05}{m}$, and the weight parameter $\rho$ is set to $1$ for IT/BT, and $v=\SI{0.15}{m}$ and $\rho=1/|\mb{P}-\mb{O_r}|^2$ for Wangsimni.
Our proposed system successfully reconstructs the 3D maps of both challenging indoor and outdoor environments.
%


\section{CONCLUSIONS}

In this paper, we propose an integrated omnidirectional localization and dense mapping system for a wide-baseline multi-camera rig with wide FOV fisheye lenses.
The proposed light-weighted deep neural networks estimate omnidirectional depth maps faster and more accurately while having a smaller number of parameters.
The output depth map is then integrated into the visual odometry, and our proposed visual SLAM module achieves better performances of pose estimation.
The extensive experiments demonstrate that our system can generate well-reconstructed 3D maps of both synthetic and real-world environments.

\small{
\section*{Acknowledgement}

This research was supported by Next-Generation Information Computing Development program through National Research Foundation of Korea (NRF) funded by the Ministry of Science, ICT (NRF-2017M3C4A7069369), and the NRF grant funded by the Korea government(MSIT)(NRF-2019R1A4A1029800).

}
\bibliographystyle{IEEEtran}
\bibliography{IEEEexample}

\end{document}